\documentclass{bmvc2k}
\usepackage[symbol]{footmisc}

\title{One-shot domain adaptation for semantic face editing of real world images using StyleALAE}

\addauthor{Ravi Kiran Reddy*}{ravikiran.reddy@iiitb.ac.in}{1}
\addauthor{Kumar Shubham*}{kumar.shubham@iiitb.ac.in}{1}
\addauthor{Gopalakrishnan Venkatesh}{gopalakrishnan.v@iiitb.ac.in}{1}
\addauthor{Sriram Gandikota}{sai.sriram@iiitb.ac.in}{1}
\addauthor{Sarthak Khoche}{sarthak.khoche@iiitb.ac.in}{1}
\addauthor{Dinesh Babu Jayagopi }{jdinesh@iiitb.ac.in}{1}
\addauthor{Gopalakrishnan Srinivasaraghavan}{gsr@iiitb.ac.in}{1}

\addinstitution{
  International Institute of Information Technology Bangalore\\
  Bangalore\\
  India
}

\runninghead{}{}

\begin{document}

\maketitle

\begin{abstract}

Semantic face editing of real world facial images is an important application of generative models. Recently, multiple works have explored possible techniques to generate such modifications using the latent structure of pre-trained GAN models. However, such approaches often require training an encoder network and that is typically a time-consuming and resource intensive process. A possible alternative to such a GAN-based architecture can be styleALAE, a latent-space based autoencoder that can generate photo-realistic images of high quality. Unfortunately, the reconstructed image in styleALAE does not preserve the identity of the input facial image. This limits the application of styleALAE for semantic face editing of images with known identities. In our work, we use a recent advancement in one-shot domain adaptation to address this problem. Our work ensures that the identity of the reconstructed image is the same as the given input image. We further generate semantic modifications over the reconstructed image by using the latent space of the pre-trained styleALAE model. Results show that our approach can generate semantic modifications on any real world facial image while preserving the identity.

\end{abstract}

\section{Introduction}
\label{sec:intro}
Semantic modifications over a human face have multiple applications including forensic art \cite{fu2010age} and cross-face verification \cite{park2010age}. Recently, many generative models have shown their efficiency in generating photorealistic human faces with required semantic modification. For example, Generative adversarial networks (GANs) like progressiveGAN \cite{karras2017progressive}  and styleGAN \cite{karras2019style} can generate high quality images with controls to create semantic modifications over attributes such as age, gender and smile \cite{shen2020interpreting}.   Despite the capability to generate semantic changes over an image sampled from the latent space, these models do not provide any encoding scheme to map a real world image onto the latent space and generate similar modifications.  Recently, \cite{zhu2020domain} proposed a two-way encoding scheme to use the pre-trained styleGAN to generate similar semantic changes over a real world human face.  Their method involves first learning an initial latent vector generated by an encoder network and then using the learned latent vector for domain regularized search.  However, their method involves training an additional encoder network that can be a time-consuming and resource intensive process.
An alternative to such GAN-based architectures are latent space-based autoencoders like styleALAE \cite{pidhorskyi2020adversarial}. Unlike other GAN-based models, styleALAE trains both an encoder network (to map real world input images to latent space) and a decoder network (to map latent vectors to image space) within the same training paradigm. This ensures that there is no additional overhead of training a separate network to work with out-of-training-dataset images for semantic manipulation. Unlike traditional autoencoders, which use MSE based reconstruction loss for image generation \cite{kingma2013auto},  styleALAE uses adversarial loss with a compositional network architecture to ensure that the generated output is of high quality. Even the decoder network in styleALAE uses the styleGAN based architecture to ensure photorealism of the generated output. It further uses latent space-based
 MSE loss to ensure reciprocity between the input image and reconstructed output image \cite{ulyanov2018takes}.
 \cite{zhu2020domain} showed that for efficient GAN-based inversion and to generate necessary semantic modifications, the encoder network needs to be trained with the gradients generated by the GAN model. Therefore, the process of training both encoder and decoder networks simultaneously in styleALAE ensures an efficient training of encoder networks using the gradients generated by the styleGAN based decoder network. At the same time, such a training process ensures that the encoder also plays its role in modeling the shared latent space. However, despite all its advantages, the reconstructed output in styleALAE i.e., D(E(I)), where D denotes decoder and E denotes encoder, does not preserve the identity of the input image (I). This limits the application of styleALAE for downstream tasks, which often consider the identity of the input image as an essential feature for semantic face editing. 

The difference in the identity of the reconstructed image and the input image is because of the disparity in the distribution learned by the model and the input data. Unlike other autoencoder-based architectures, styleALAE uses latent space-based MSE loss to ensure the reciprocity between the encoded image and the reconstructed output \cite{ulyanov2018takes, srivastava2017veegan}. Even though, this ensures that the decoder network captures multiple semantics from the input image, it fails to ensure the identity-based similarity between the reconstructed output and the input image. Recently, \cite{yang2020one} proposed a framework to generate images from the same distribution as that of a given one-shot example in styleGAN based model. The complete process consists of two steps i.e. the iterative latent optimization \cite{abdal2019image2stylegan} to project input image onto the latent space of pre-trained styleGAN and followed by fine-tuning the weights of the styleGAN while fixing the projected latent vector to match the perceptual and pixel-wise similarity with the given input image. This one-shot training process ensures that the manifold structures of styleGAN shift toward the target distribution and can further be reused to generate multiple images with similar identity as the input image. A similar trick to preserve the identity has been proposed in \cite{zakharov2019few} and \cite{kowalski2020config}, where one-shot fine-tuning of the generator reduces the identity difference in deepfake and computer graphics-based semantic face editing system.

In our work, we propose a one-shot domain adaptation method to solve the mismatch between the identity of the input image and the reconstructed output of styleALAE.  We ensure this by first mapping the input image onto the latent distribution of pre-trained styleALAE, and then finetuning the decoder to produce the required manifold shifting toward the given input image. We further show that the latent space of the pre-trained styleALAE model can be reused to generate age, smile, and gender-based semantic modifications under the given one-shot domain adaptation technique. Unlike the deepfake and graphics-based face editing methods, which use conditional GAN to generate semantic modification, our approach uses a latent space-based traversal scheme \cite{shen2020interpreting} to generate the required semantic modification on the input image. We further experiment with different inversion techniques to project the input image onto the latent space of styleALAE and show its effect on the overall semantic modification and identity preservation. 
Our contribution can be summarized as follow. 
\begin{itemize}
\item  We propose a one-shot domain adaption technique to solve the disparity in the identity of the input image and reconstructed output in styleALAE.
\item  We experiment with different inversion techniques and show their impact on the semantic modification and identity preservation in styleALAE.
\item We show that within one-shot domain adaptation technique, the latent space of pre-trained styleALAE can be reused to generate semantic modification using a linear traversal-based scheme.
\end{itemize}

\section{Related Work}
\subsection{Generative models}
Recently, a lot of progress has been made in the area of semantic face editing. Models like in \cite{kingma2018glow} have shown an efficient method for semantic face editing of real world images by learning a disentangled latent structure. However, the high dimension of the latent space of such a network often makes it cumbersome to train. Researchers have also explored conditional GAN-based model \cite{he2019attgan,lu2018attribute,gu2019mask} for semantic face editing, but training such a model often requires real world images with necessary semantic variations, which is often hard to obtain. To particularly tackle the problem of lack of dataset \cite{kowalski2020config} used graphic rendering engines to generate synthetic face images. They then trained an autoencoder with a shared latent space between synthetic dataset and real world images. They further used this dataset to generate semantic modification over the input image. However, in such an approach the semantic modification over in any input image is often tied up with the variation captured in the synthetic dataset.

\subsection{GAN Inversion}
With the recent success of models like progressiveGAN \cite{karras2017progressive}, styleGAN \cite{karras2019style} and StyleGAN2 \cite{karras2020analyzing} in generating photorealistic images, researchers have focused on enhancing its application in downstream face editing tasks. \cite{shen2020interpreting} showed that the latent structure of these models have a defined semantic structure, and a linear traversal along a specific direction can be utilized for semantic face editing. Similarly, \cite{abdal2020styleflow,shubham2020learning} have explored a nonlinear traversal scheme for face editing. To extend their application on real world images with known identities, \cite{abdal2019image2stylegan} and \cite{zhu2020domain} proposed a latent space-based optimization method to map any input image onto the latent space of the existing pre-trained model. \cite{zhu2020domain} further showed that the latent space of styleGAN can be reused to generate required semantic modification over the mapped image. However, their approach involves training an additional encoder network that provides a domain regularization for latent optimization. This puts an additional overhead of training a separate network, particularly for real world images. 

\subsection{Few-shot domain adaptation} 

Training a model to generate photo-realistic and semantically rich facial images often requires a large dataset. Recently, there has been a growing interest in generalizing the performance of deep learning models with few sets of images \cite{zakharov2019few,motiian2017few,liu2019few,finn2017model}. Multiple techniques have been proposed to achieve the given task. \cite{finn2017model} proposed meta-learning approaches to adapt the model for any new task. Similarly, \cite{snell2017prototypical,vinyals2016matching} focused on learning the embedding space better suited for few-shot learning. Recently \cite{yang2020one} proposed a one-shot domain adaptation framework for generating images with different styles using pre-trained styleGAN. The authors first mapped the given input image onto the latent space of styleGAN and then fine-tuned the generator to produce the necessary manifold shifting for required face editing. One-shot domain adaptation techniques have also been used for identity preservation. \cite{zakharov2019few} and \cite{kowalski2020config} showed that the disparity in the identity of the reconstructed output for a given input image can be mitigated by fine-tuning the generator using one-shot training. 

\section{Method}
Unlike other works in GAN inversion, ours utilizes the encoder network of the pre-trained styleALAE model to map real world images onto the latent space. We further explore the properties of the encoder network to generate semantic modifications over images using a one-shot domain adaptation technique.  Such approaches have previously been studied only for style transfer and identity preservation. Similarly, unlike other generative models, which require synthetic datasets for semantic modifications, ours utilized the existing latent space of styleALAE to generate similar changes on the reconstructed images of real-world facial images.\\

Our algorithm therefore consists of the following three crucial steps. 
\begin{itemize}
\item \textbf{Image inversion}:  For any given input image, a corresponding latent vector is generated by projecting the image onto the latent space of pre-trained styleALAE model. In our work, we have experimented with different projection schemes for image inversion and have shown its impact on the overall semantic modification (see Sec \ref{sec:inversion} and Sec \ref{sec:exp}  ). 

\item \textbf{Manifold shifting}: Once the latent vector is generated for the given input image, we fix the corresponding latent vector and update the weights of the decoder of styleALAE model to minimize the difference between the input image and the reconstructed output image. This optimization generates the manifold shift \cite{yang2020one} in the output distribution towards the given input image. 

\item \textbf{Semantic modification}: Subsequently, we use the linear traversal scheme on the latent space of styleALAE to generate the necessary semantic modification \cite{shen2020interpreting}. 
\end{itemize}

\subsection{Image Inversion} 
\label{sec:inversion}
The first step of the one-shot domain adaptation method \cite{yang2020one} involves projecting the input image onto the latent space of the styleALAE. While other GAN-based approaches do not provide any encoder network to map the input image onto the latent space, styleALAE on the other hand provides an encoder to project them onto the learned latent space.

Recent works in GAN inversion \cite{zhu2020domain,abdal2019image2stylegan} have shown that the styleGAN has a latent vector associated with images of people from the real world with known identities. This latent vector, when passed through the generator of the original styleGAN, can generate a perceptually similar image with the same identity as the input image. \cite{zhu2020domain} have further used this learned vector for semantic modification, and \cite{yang2020one} have shown its suitability for domain adaptation purposes. However, we noticed that such approaches, when applied to the latent space styleALAE, do not generate a high quality image and sometimes even fail to generalize for different identities (Section \ref{sec:exp}).

In our experiments, we found that the latent vector generated using the pre-trained encoder network in styleALAE is more suited for projecting images for one-shot domain adaptation, particularly for subsequent semantic modification (Section \ref{sec:semantic_mod}). The encoder of the styleALAE network is trained with the gradients of the generator and is efficient in capturing the high level semantics features such as color, race, gender and age. This helps the encoder network in projecting any input image closer to the learned manifold structure which can later be used to generate semantic modifications (Section \ref{sec:semantic_mod}).

\subsection{Manifold shifting}
\label{sec:manifold_shift}
 After generating the latent vector associated with the given input image, the next step is to generate a manifold shift in styleALAE towards the given input image \cite{yang2020one}. For this,  we finetune the decoder of styleALAE using pixel and perceptual similarity while fixing the projected latent vector during the complete fine tuning process. The finetuning  of the decoder reduces the identity gap between the input image, and the reconstructed image associated with the latent vector generated by the encoder network. The change in the identity of the reconstructed image is not only visible for the projected latent vector, but in fact, using the finetuned decoder, even the neighborhood of the projected vector in the latent space generates images that resemble the identity of the given input image.

In our work, we finetune the decoder of styleALAE using a weighted combination of VGG-16 based perception loss \cite{abdal2019image2stylegan} and pixel wise MSE loss.

\begin{equation}
    \label{eqn:total_loss}
    \mathcal{L}_{ft} =  \lambda_{MSE}|| I - D(E(I)) || ^{2} + \lambda_{vgg} L_{percep}(D(E(I)), I) 
\end{equation}

where $I \in {\rm I\!R}^{n x n x 3}$ and $D(.)$, $E(.)$  are the decoder and encoder of styleALAE.

For perception loss in Equation \ref{eqn:total_loss} we have used smooth L1 loss \cite{rashid2017interspecies} over VGG16 features:
\begin{equation}
    \begin{split}
     & L_{percep}(I_1, I_2) = \sum_{j=1}^4 z_j \text{  where,  } \\
     & z_j =
     \begin{cases}
        0.5||F_j(I_1) - F_j(I_2)||^2 & if ||F_j(I_1) - F_j(I_2)|| < 1 \\
        ||F_j(I_1) - F_j(I_2)|| - 0.5 & otherwise 
     \end{cases}
    \end{split}
\end{equation}
 $I_1, I_2 \in {\rm I\!R}^{n x n x 3}$ are images used for comparison and $F_j$ are VGG16 features from \textit{conv}1\_1,  \textit{conv}1\_2, \textit{conv}3\_2 and \textit{conv}4\_2 respectively \cite{abdal2019image2stylegan}. 
Perceptual loss in the overall loss function, encourages the model to generate similar feature representation between input image and the reconstructed output, and act as a regularizer to ensure smooth optimization \cite{abdal2019image2stylegan}

\subsection{Semantic Modification}
\label{sec:semantic_mod}
Once the required manifold shift has been generated in styleALAE toward given input image, we reuse the latent space of the pre-trained model to generate semantic modifications. For this, we modify the projected latent vector using linear interpolation along the hyper-plane trained to segregate latent vectors with semantically opposite features \cite{shen2020interpreting}.
\begin{equation}
    \mathcal{L}_{edit} = \mathcal{L}_{projected} + \alpha\mathcal{N}
\end{equation}
where, $\mathcal{L}_{edit}$ is the new edited latent vector, $\mathcal{L}_{projected}$ is the original projected vector, $\mathcal{N}$ is the hyper-plane for the required semantic modification and $\alpha$ is a hyperparameter. 
\section{Experiments}
\label{sec:exp}
For our experiments, we use a styleALAE model pre-trained on 1024 $\times$ 1024 $\times$ 3 images from the FFHQ dataset. We have compared our formulation (one-shot adaptation + encoder based projection) with four different algorithms namely \textbf{(i)} vanilla styleALAE \textbf{(ii)} latent optimization based inversion techniques \cite{abdal2019image2stylegan} where the projection vector generated in styleALAE by the encoder network is optimized using the pixel and perceptual based loss (see Equation \ref{eqn:total_loss}); this involves only latent optimization. \textbf{(iii)} one-shot domain adaptation in styleALAE with randomly sampled vector as projection vector; this involves one-shot adaptation and random projection. In this method the decoder of styleALAE model is fine-tuned by fixing a randomly sampled latent vector as projection vector during decoder based fine tuning and then the same vector is later used for semantic modification. \textbf{(iv)} one-shot domain adaptation in styleALAE with latent optimization-based projection \cite{yang2020one} (involving one-shot adaptation as well as latent optimization). In this method the encoded latent vector is first optimized using latent optimization techniques and later fixed during the complete fine tuning of the decoder. The generated latent vector is then reused for semantic modifications. To generate semantic modifications, we have used the default latent directions provided by the authors of styleALAE for all the above mentioned algorithms. 
The experiments were performed on a Intel(R) Xeon Silver 4114 Processor @ 2.20GHz with 40 cores and 128GB RAM, we have used one Nvidia GeForce RTX 2080Ti with 11016 MB VRAM.

\subsection {Quantitative evaluation}
To compare the performance of our proposed method on real world identities, we have used 1000 single face images of celebrities from IMDB-wiki face dataset \cite{Rothe-IJCV-2018,Rothe-ICCVW-2015}. In our quantitative evaluation, we have done two kinds of analyses. First, we compare the input image with the reconstructed output generated after the manifold shift (Section \ref{sec:manifold_shift}) and second, we compare the images generated during linear traversal for semantic modification (Section \ref{sec:semantic_mod}) with the given input image. 

As a metric, we have used SSIM, PSNR \cite{hore2010image} and SWD \cite {rabin2011wasserstein} scores, to compare different algorithms. 
\begin{table}
    \centering
     \resizebox{0.8\columnwidth}{!}{
    \begin{tabular}{|c|c|c|c|c|}
    \hline
     Algorithm & SSIM $\uparrow$ & PSNR $\uparrow$ & SWD $\downarrow$  \\
    \hline
    Vanilla styleALAE& 0.596 / 0.072 & 15.823 / 2.029 & 1839.956 / 156.494  \\
    Only latent optimization & 0.746 / 0.063  & 22.759 / 1.170 & 1489.477 / 164.251 \\
    one-shot adaptation + random projection & 0.853 / 0.040 & 27.461 / 1.903 & 1194.786 / 155.150 \\
    one-shot adaptation + latent optimization & \textbf{0.887 / 0.037} & \textbf{29.407 / 2.152} & \textbf{1079.925 / 163.816} \\
    one-shot adaptation + encoder based projection (Ours) & 0.881 / 0.039 & 29.061 / 2.197 & 1098.163 / 168.272 \\
    \hline     
    \end{tabular}
    }
    \caption{Quantitative comparison between the reconstructed output and input image for different methods.}
    \label{tab:inv_comp}
\end{table}

Table \ref{tab:inv_comp} shows the comparison score for the reconstructed image and the input image. Among different methods ((i), (ii), (iii), (iv)) the one-shot domain adaptation method with latent optimization based projection performs best, this is closely followed by our method. This high
generation quality of reconstructed output in latent optimization based projection and encoder based projection when compared to other methods can be attributed to better initialization provided by these methods for fine-tuning of decoder.

\begin{table}[h!]
    \centering
    \resizebox{0.8\columnwidth}{!}{
    \begin{tabular}{|c|c|c|c|c|c|}
    \hline
    Type & Algorithm & SSIM $\uparrow$ & PSNR $\uparrow$ & SWD $\downarrow$  \\
    \hline
     & Vanilla styleALAE & 0.567 / 0.071 & 15.052 / 1.817 & 1974.962 / 140.477 \\
     AGE & Only latent optimization & 0.591 / 0.069 & 16.020 / 1.803 & 1912.731 / 141.372  \\
     & one-shot adaptation + random projection & 0.692 / 0.046 & 18.948 / 1.259 & 1786.257 / 114.478 \\
     & one-shot adaptation + latent optimization & 0.684 / 0.061 & 18.135 / 1.685 & 1711.060 / 132.413  \\
     & one-shot adaptation + encoder based projection (Ours) & \textbf{0.787 / 0.049} & \textbf{22.715 / 1.670} & \textbf{1490.155 / 147.353}  \\
    
     \hline
     & Vanilla styleALAE & 0.581 / 0.071 & 15.530 / 1.950 & 1874.735 / 139.257 \\
     Gender & Only latent optimization & 0.6073 / 0.069 & 16.625 / 1.956 & 1805.526 / 139.186  \\
     & one-shot adaptation + random projection & 0.726 / 0.048 & 20.521 / 1.596 & 1602.929 / 122.529 \\
     & one-shot adaptation + latent optimization & 0.708 / 0.060 & 19.009 / 1.822 &  1606.178 / 130.274  \\
     & one-shot adaptation + encoder based projection (Ours) & \textbf{0.820 / 0.045} & \textbf{24.870 / 1.882} & \textbf{1330.361 / 141.232}  \\
     
     \hline
      & Vanilla styleALAE & 0.566 / 0.071 & 15.415 / 1.897 & 1886.499 / 141.816 \\
     Smile & Only latent optimization & 0.592 / 0.068 & 16.500 / 1.892 & 1816.944 / 142.755  \\
     & one-shot adaptation + random projection & 0.723 / 0.049 & 20.987 / 1.366 & 1589.207 / 131.564 \\
     & one-shot adaptation + latent optimization & 0.643 / 0.061 & 17.927 / 1.613 & 1695.687 / 127.006  \\
     & one-shot adaptation + encoder based projection (Ours) & \textbf{0.786 / 0.046} & \textbf{23.831 / 1.559} &\textbf{1398.035 / 149.570}  \\
     
    \hline
    \end{tabular}
    }
    \caption{Quantitative comparison between different methods for semantic modification.}
    \label{tab:semantic_comp}
\end{table}

Table \ref{tab:semantic_comp} shows the comparison between semantically modified images with the input image. As per the result it is evident that our method generates high quality images with greater perceptual similarity. This can be attributed to the ability of the encoder network to capture multiple semantic features of input image and generate a latent vector which lies closer to the true latent manifold structure of styleALAE.

\subsection{Qualitative evaluation}
\begin{figure}[h!]
    \centering
    \includegraphics[width=0.52\linewidth]{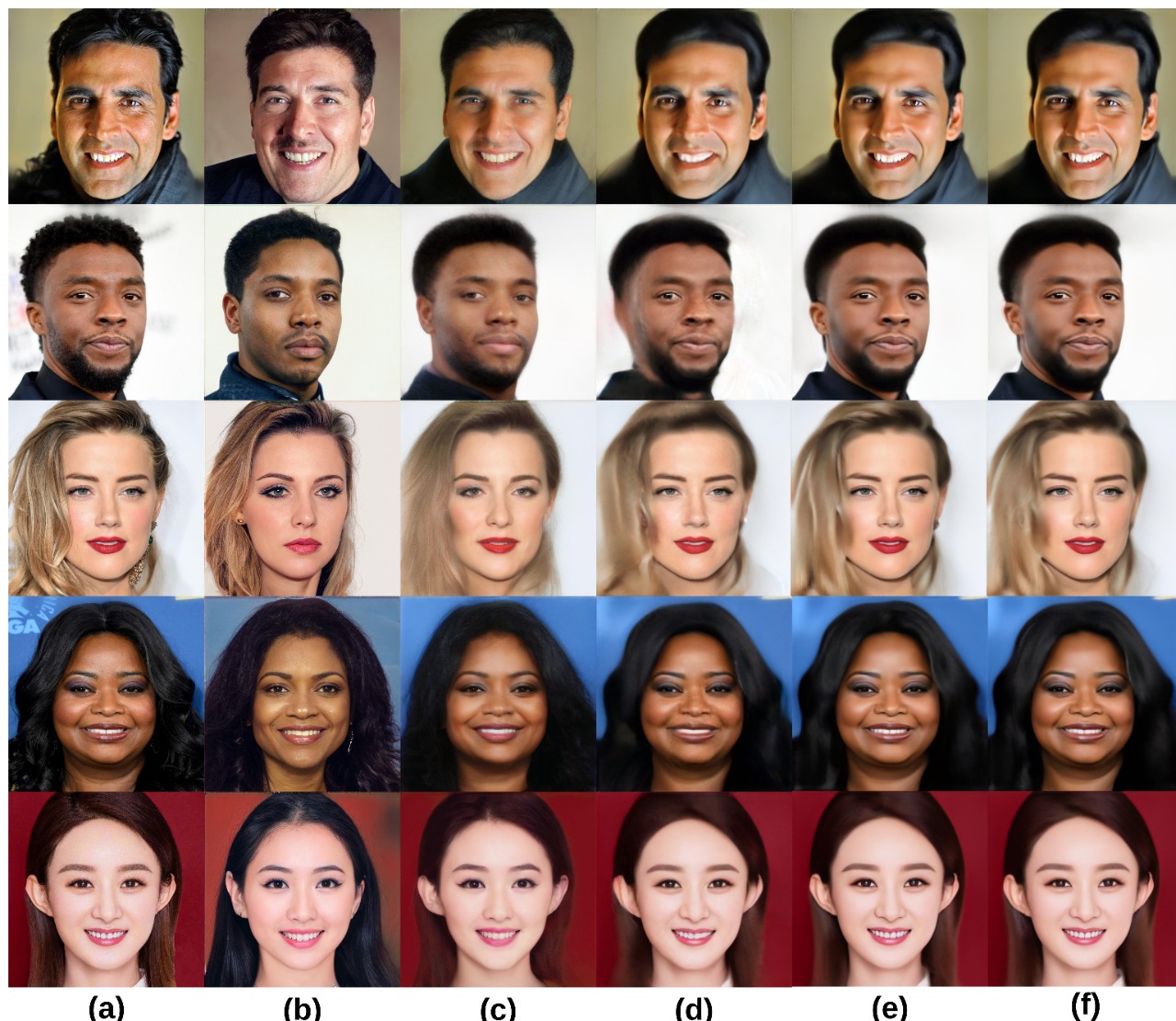}
    \caption{Qualitative comparison on image reconstruction with different approaches. (a) Original image used for image inversion (b) vanilla styleALAE (c) only latent optimization (d) one-shot adaptation + random projection (e) one-shot adaptation + latent optimization (f) one-shot adaptation + encoder based projection (Ours)  }
    \label{fig:qual_inv}
\end{figure}

Figure - \ref{fig:qual_inv} shows the qualitative comparison between reconstructed output and the input image for all the algorithms. As can be seen, when compared to the vanilla styleALAE or only latent optimization based methods, one-shot domain adaptation does a better job in preserving the identity of the reconstructed image irrespective of the choice of the projection vector. 

\begin{figure}[h!]
    \centering
    \includegraphics[width=0.52\linewidth]{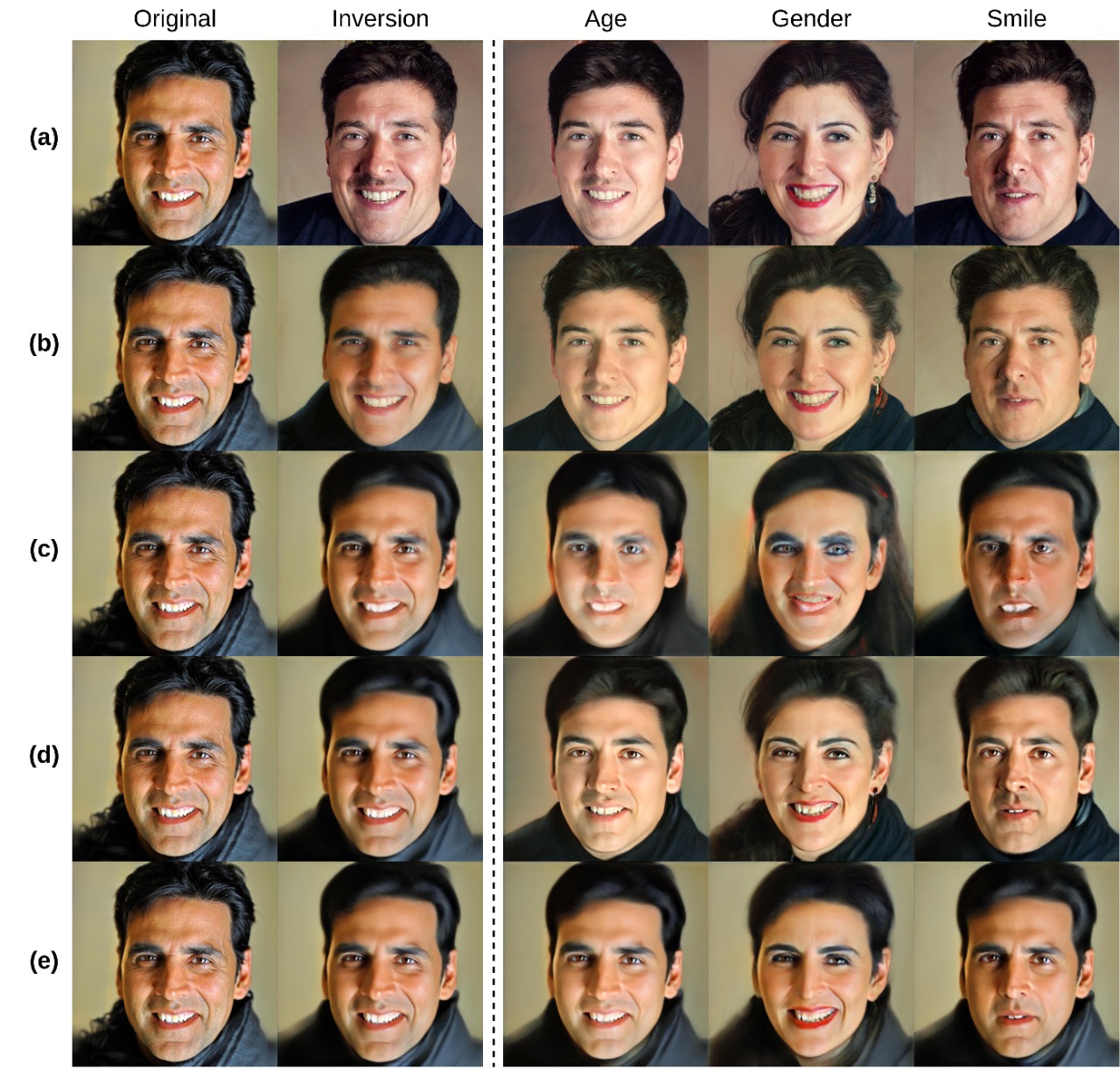}
    \caption{Qualitative comparison of different methods for age and smile based semantic manipulation over the reconstructed image. (a) vanilla styleALAE (b) only latent optimization (c) one-shot adaptation + random projection (d) one-shot adaptation + latent optimization (e) one-shot adaptation + encoder based projection (Ours) }
    \label{fig:semantic_comp}
\end{figure}

Although the impact of different projection vectors for one-shot domain adaptation is much more evident in semantic modifications. As can be seen in Figure \ref{fig:semantic_comp},  a random projection vector generates unwanted artifacts on the input image, while latent optimization based projection vector and one-shot adaptation with latent optimization generates images with different identity. Compared to other algorithms our method fairs well in generating the necessary semantic modification while preserving the identity. 

\subsection{User study}





\begin{figure}[h!]
    \centering
    \includegraphics[width=0.45\linewidth]{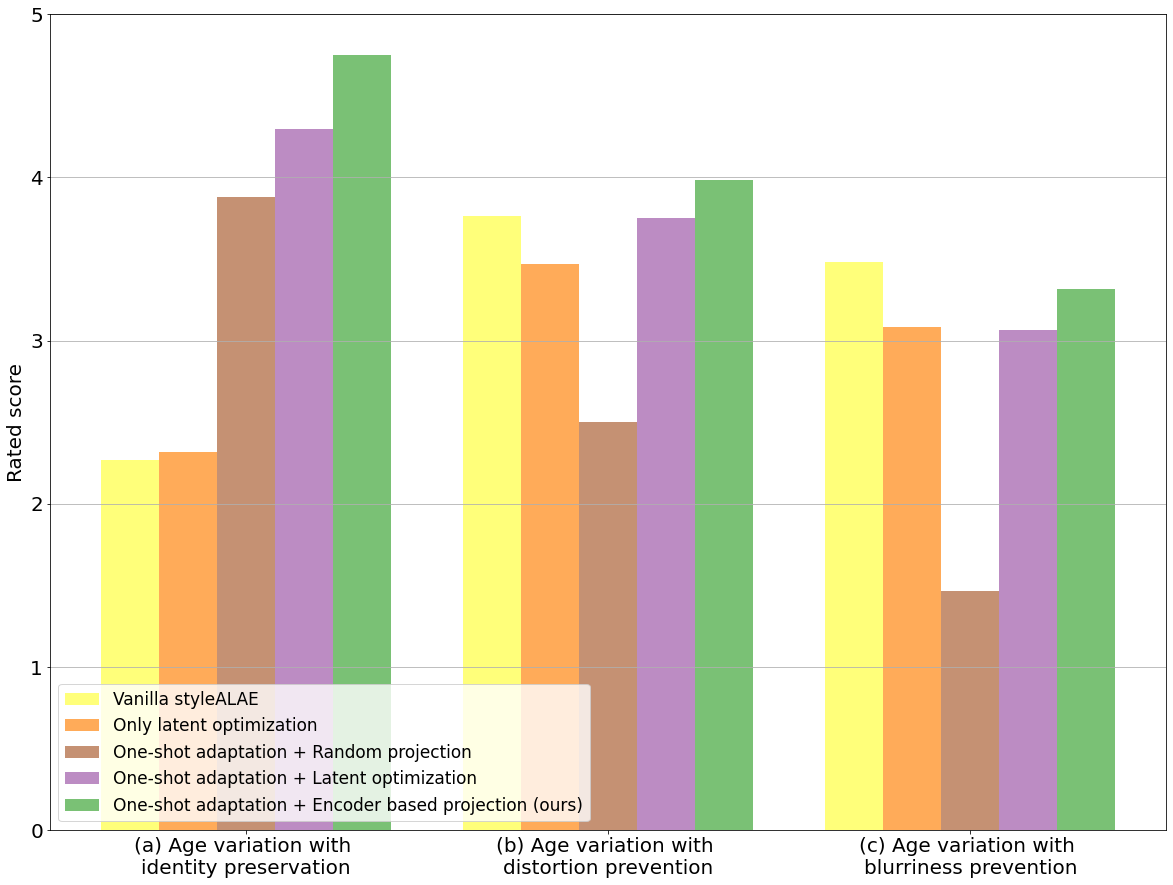}\hfill
    \includegraphics[width=0.45\linewidth]{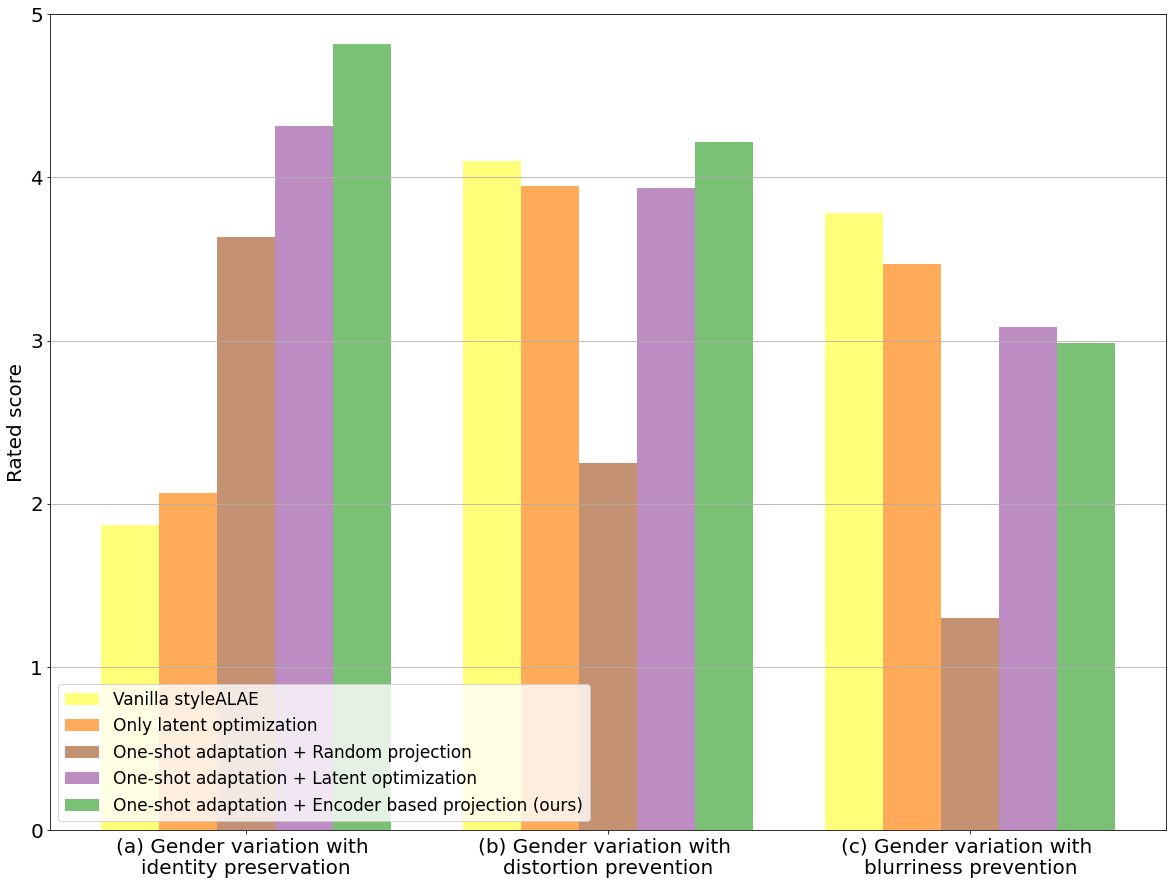}
    \includegraphics[width=0.45\linewidth]{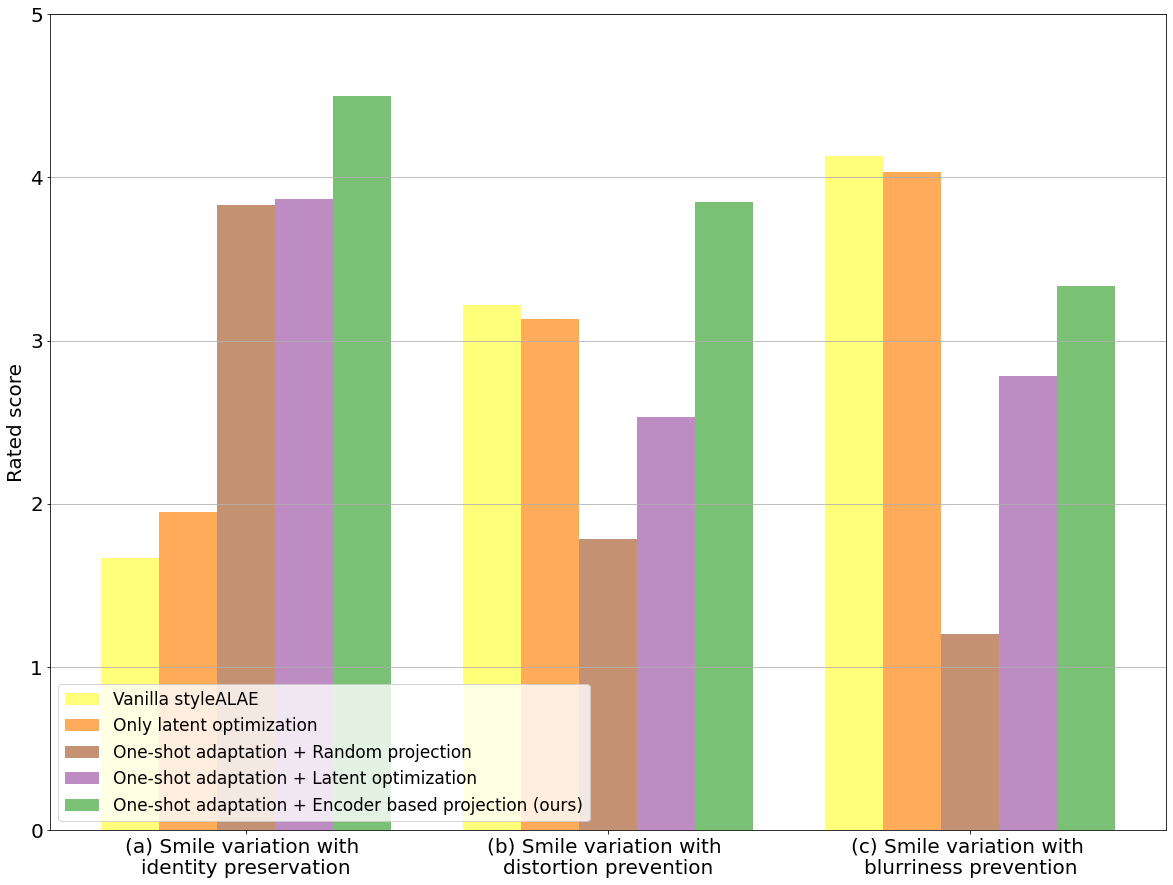}
    \caption{User ratings for the semantic face editing.}
    \label{fig:user_study}
\end{figure}


Considering the subjective nature of perceiving semantic manipulations and identity preservation, we have further evaluated the algorithms using human evaluators. For this, 450 different trajectories of semantic modifications were generated and verified by two annotators on a 5 point likert scale. 
The evaluators rated the algorithms on following dimensions  \textbf{a}: Rate the algorithms on the basis of their ability to preserve the identity of the input image while generating the required semantic manipulation. \textbf{b}: Rate the algorithms on the basis of their ability to prevent image distortion  while generating the required semantic manipulation. \textbf{c}: Rate the algorithms on the basis of their ability to prevent blurring while generating the required semantic manipulation.

Figure \ref{fig:user_study} shows the results of the user study for the mentioned algorithms for age, gender and smile based manipulation. From the results it is evident that our approach outperforms other methods in terms of identity preservation and prevention of image distortion. 

As per the results (see Figure \ref{fig:semantic_comp} and Figure \ref{fig:user_study} ), it is evident that the latent vector generated by the encoder network of styleALAE does a better job in preventing image distortion and preserving the identity of the input image in a linear traversal based semantic modification scheme. As shown in Figure  \ref{fig:qual_inv}, the finetuning of the decoder in styleALAE ensures that the reconstructed output preserves the identity of the input image irrespective of the choice of the image inversion technique. But, in a linear traversal based semantic modification scheme, it is essential to be closer to the underlying manifold structure of styleALAE to generate semantic modifications. A random vector or a latent optimization-based projection vector fails to generate a vector closer to the existing manifold structure of styleALAE. It hence performs poorly in preserving identity or avoiding any distortion in the generated output. On the other hand, the encoder network ensures that the projected latent vector is generated closer to the underlying manifold structure and hence generates better output.

\section{Conclusion}
In our work, we have addressed the problem of difference in identity of the reconstructed output compared to given input image in styleALAE, and its impact on generating the semantic modification over real world images.
 Our results shows that with one-shot domain adaptation, the identity of input image can be preserved and under the given setting, the latent space of styleALAE can be reused to generate semantic modification over the given input image with known identity. We have further shown the importance of the projected vector in one-shot domain adaptation for latent space based semantic modification scheme and in generating high quality images.  In the future, we will extend this formulation to other attributes. We also intend to further investigate the impact of fine tuning of the generator on flow-based and auto-regressive models.   

\bibliography{egbib}
\end{document}